\documentclass[letterpaper]{article} 
\usepackage[submission]{aaai2026}  
\usepackage{times}  
\usepackage{helvet}  
\usepackage{courier}  
\usepackage[hyphens]{url}  
\usepackage{graphicx} 
\usepackage{natbib}  
\usepackage{caption} 
\usepackage{algorithm}
\usepackage{algorithmic}
\usepackage{amsmath}
\usepackage{amsfonts}
\usepackage{amsthm}   
\usepackage{booktabs}
\usepackage{subcaption}
\usepackage{newfloat}
\usepackage{listings}
\DeclareCaptionStyle{ruled}{labelfont=normalfont,labelsep=colon,strut=off} 
\floatstyle{ruled}
\newfloat{listing}{tb}{lst}{}
\floatname{listing}{Listing}
\newtheorem*{assumption*}{\assumptionnumber}
\providecommand{\assumptionnumber}{}
\makeatletter
\newenvironment{assumption}[2]
 {%
  \renewcommand{\assumptionnumber}{Assumption #1-$\mathcal{#2}$}%
  \begin{assumption*}%
  \protected@edef\@currentlabel{#1-$\mathcal{#2}$}%
 }
 {%
  \end{assumption*}
 }
\makeatother

\newtheorem{theorem}{Theorem}
\newtheorem{corollary}{Corollary}[theorem]
\newtheorem{lemma}[theorem]{Lemma}

\newtheorem{definition}{Definition}

\title{Low-Bit, High-Fidelity: Optimal Transport Quantization for Flow Matching}

\author{
    Dara Varam\textsuperscript{\rm 1,2},
    Diaa A. Abuhani\textsuperscript{\rm 2},
    Imran A. Zualkernan\textsuperscript{\rm 1},
    Raghad AlDamani\textsuperscript{\rm 1},
    Lujain Khalil\textsuperscript{\rm 1}
}

\affiliations{
    \textsuperscript{\rm 1} Department of Computer Science \& Engineering, 
    American University of Sharjah, Sharjah, UAE\\\texttt{\{b00081313, izualkernan, g00085944, g00082632\}@aus.edu}\\[0.75em]
    \textsuperscript{\rm 2} Senseable City Laboratory, Massachusetts Institute of Technology, 
    Cambridge, Boston, USA\\
    \texttt{\{varam, diaa\}@mit.edu}
}

\usepackage{bibentry}
\makeatletter
\def\showauthors@on{T}  
\makeatother

\begin{document}

\maketitle

\begin{abstract}
Flow Matching (FM)  generative models \cite{FM} offer efficient simulation-free training and deterministic sampling, but their practical deployment is challenged by high-precision parameter requirements. We adapt optimal transport (OT)-based post-training quantization to FM models, minimizing the 2-Wasserstein distance between quantized and original weights, and systematically compare its effectiveness against uniform, piecewise, and logarithmic quantization schemes. Our theoretical analysis provides upper bounds on generative degradation under quantization, and empirical results across five benchmark datasets of varying complexity show that OT-based quantization preserves both visual generation quality and latent space stability down to 2–3 bits per parameter, where alternative methods fail. This establishes OT-based quantization as a principled, effective approach to compress FM generative models for edge and embedded AI applications\footnote{Code for reproducibility and validation can be accessed through https://github.com/Quantized-FM/LowBit-HighFid.}.
\end{abstract}


\section{Introduction}
Flow Matching (FM) generative models have emerged as a powerful paradigm for efficient, high-fidelity generation. Recent works highlight their potential as a robust and stable alternative to diffusion models, owing to their simulation-free training and deterministic sampling properties \cite{lipman2023flow, shih2023parallel, zhang2023diffcollage}. However, like other generative models, FM architectures remain challenging to deploy on resource-constrained devices due to their reliance on high-precision parameters, making effective quantization a critical open problem  \cite{fortuin2023challenges, manduchi2024challenges}. Consequently, understanding how to compress these models aggressively while maintaining generation quality is important for expanding the practical utility of FM models in edge and embedded settings.

As the need to deploy generative models on edge devices grows, quantization has become a widely explored approach for both generative and discriminative deep learning architectures. Recent works have demonstrated success in quantizing diffusion models using specialized techniques such as Q-Diffusion \cite{li2023q} and BiDM \cite{NEURIPS2024_44b61c5c}. These methods build on earlier attempts to quantize Generative Adversarial Networks (GANs), such as QGANs \cite{wang2019qganquantizedgenerativeadversarial}. However, such methods are tailored for specific classes of generative models, and adapting them to ODE-based architectures such as FM models is nontrivial due to differences in vector field parameterization, loss functions, and sampling paths \cite{draxler2024free, kobyzev2020normalizing}. This complexity is further compounded by the fact that in FM models, small quantization errors in the weights can propagate nonlinearly through the integration steps \cite{zhou2025an}. As a result, the generated samples may exhibit disproportionately large deviations, often exceeding those observed in diffusion models. 

\begin{figure*}
    \centering
    \includegraphics[width=0.9\linewidth]{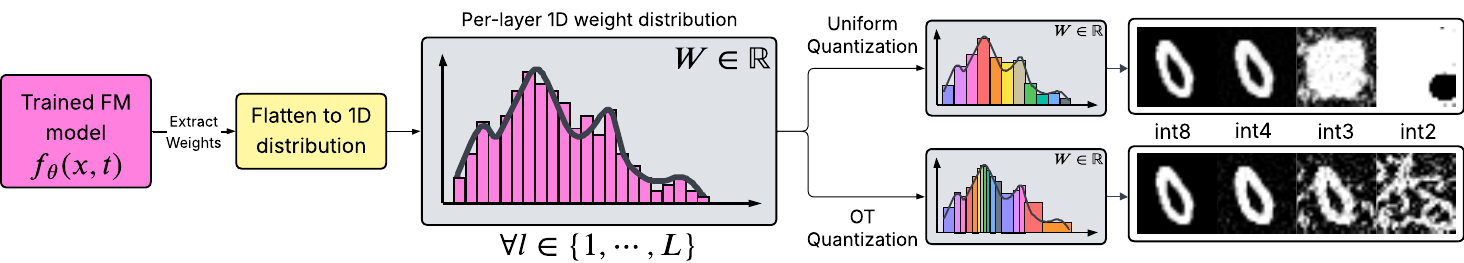}
    \caption{System diagram illustrating the quantization process for Flow Matching (FM) generative models. Per-layer weights are extracted from a trained FM model and flattened into a one-dimensional distribution. These weights are quantized using optimal transport (equal-mass bin) quantization method. Quantized models with different bit-widths are then used for sample generation, and the resulting outputs are evaluated for generative fidelity while the latent space is used to evaluate disentanglement.}
    \label{fig:system-diagram}
\end{figure*}

In this work, we systematically investigate the influence of optimal transport (OT)-based post-training quantization on FM generative models, with a particular focus on extreme low-bit regimes. While OT-based quantization has shown promise in other contexts \cite{KREITMEIER20111225}, we believe that its impact on the generative fidelity and latent space disentanglement of FM models has not been thoroughly studied. Our empirical and theoretical analyses reveal that OT quantization can preserve both sample quality and latent structure even at aggressive compression levels, enabling practical deployment of FM models on edge devices. These findings provide valuable insights for the efficient compression and deployment of advanced generative models in real-world, resource-constrained environments. A brief visual summary of the approach is shown in Figure \ref{fig:system-diagram}.

\section{Related Work}
Model compression via quantization has been widely explored, primarily through post-training quantization (PTQ) and quantization-aware training (QAT) \cite{abushahla}. PTQ methods in particular have shown remarkable merit in low-bit deployment of both discriminative and generative models without full retraining \cite{WAN2020107338, zeng2025diffusionmodelquantizationreview}. Early efforts quantized GANs using calibrated rounding, achieving 4-bit weights and 8-bit activations with minimal quality loss \cite{andreev2022quantization}. 

For diffusion models, methods such as NDTC \cite{shang2023post}, Q-Diffusion \cite{li2023q}, BiDM \cite{NEURIPS2024_44b61c5c}, and PTQD \cite{he2023ptqd} introduced timestep-aware calibration, block-wise quantization, and noise scheduling, achieving 1–8 bit quantization with little drop in FID or Inception Score on datasets like CIFAR-10 and ImageNet. Further, recent approaches like APQ-DM \cite{wang2024towards}, PTQ4DiT \cite{wu2024ptq4dit}, PQD \cite{ye2025pqd}, DiTAS \cite{dong2025ditas}, and Q-DiT \cite{chen2024q} have extended PTQ to transformer-based diffusion models, leveraging channel-wise activation smoothing and sample-aware scaling to enable near-lossless quantization to 4–8 bits.

However, despite the close mathematical relationship (where diffusion models can be seen as a special case of FM) these quantization techniques are fundamentally tailored to the stochastic, timestep-dependent nature of diffusion. They rely on design choices such as calibration over timesteps or denoising noise scheduling that do not directly transfer to the continuous, ODE-based formulation of FM models. FM’s reliance on deterministic integration of a learned vector field, rather than iterative denoising, means quantization errors propagate differently and may impact fidelity and latent structure \cite{NEURIPS2023_8e63972d} in ways not captured by diffusion-oriented methods. As such, there remains an open gap in understanding how quantization strategies, especially those based on OT, affect FM models specifically.

\section{Preliminaries}
Let $p_0$ be the base distribution initialized to $\mathcal{N}(0, I)$ and $f_\theta:\mathbb{R}^d \times [0,T] \rightarrow \mathbb{R}^d$ be the trained velocity network with weights $\theta \in \mathbb{R}^p$ and $T$ terminal time (typically $1$) of the probability flow ODE. We make the following assumptions:
\begin{assumption}{1}{A} \label{ass:1a}
    The trained FM velocity network $f_\theta$ with parameters $\theta$ is state-Lipschitz, or, mathematically: 

    \begin{equation}
        ||f_\theta(x,t) - f_\theta(x',t)|| \leq L_x||x-x'|| \space \forall x, x', t
    \end{equation}
\end{assumption}
\begin{assumption}{1}{B} \label{ass:1b}
    $f_\theta$ is parameter-Lipschitz under worst-case sensitivity ($\sup$-norm): 

    \begin{equation}
        ||f_{\theta + \Delta \theta}- f_\theta|| \leq L_\theta^\infty||\Delta \theta||_\infty
    \end{equation}
\end{assumption}

\begin{assumption}{1}{C} \label{ass:1c}
    $f_\theta$ is parameter-Lipschitz under root mean-squared sensitivity ($l_2$-norm): 

    \begin{equation}
        ||f_{\theta + \Delta \theta}- f_\theta|| \leq L_\theta^2||\Delta \theta||_2
    \end{equation}
\end{assumption}

\begin{assumption}{1}{D} \label{ass:1d}
    If $\phi$ represents the embeddings for Inception-v3, then $\phi$ is $L_\phi$-Lipschitz.
\end{assumption}

\begin{assumption}{1}{E} \label{ass:1e}
    After mapping images into the Inception feature space, both the quantized and standard models can be well-approximated by a multivariate Gaussian distribution. 
\end{assumption}

\ref{ass:1a} is a regularity assumption that formally states that the vector field $f_\theta (x,t)$ cannot grow faster than some scalar $L_x$ multiplied by the distance between two states $x$ and $x'$. \ref{ass:1b} and \ref{ass:1c} measure the networks' worst-case and average sensitivity to perturbations, respectively. We assume that if a weight matrix is perturbed by at most $\Delta \theta$, the vector field's perturbation is bounded by at most a controlled multiple ($L_\theta^\infty$) of $||\Delta \theta||_\infty$. A similar intuition can be used for \ref{ass:1c}. Assumption \ref{ass:1d} states that the Inception-v3 feature extractor does not magnify pixel-space differences by more than a fixed factor $L_\phi$. 

Assumption \ref{ass:1e} builds on prior work that have made similar assumptions in theoretical derivations, including \cite{heusel2017gans, lucic2018gans}. Such an assumption is necessary for the derivations that follow, but its limitations can be noted in works such as \cite{jayasumana2024rethinking, luo2024beyond, 10.1145/3708778.3708790}.

\subsection{Uniform Quantization}
For a weight matrix $W$ in a network, uniform PTQ defines a single, symmetric range characterized by $[-R, R]$ large enough to cover all weights in the layer. With $b$ bits, the step-size $\Delta$ is defined as $\Delta = \frac{2R}{2^b}$. We further define the worst-case uniform quantization error per weight $w \in W$ as $\delta_U\leq \frac{\Delta}{2} = \frac{R}{2^{b-1}}$, reflecting the maximum perturbation that a quantized weight value (in a weight matrix) can have from the true value. Under Assumption \ref{ass:1b}, it follows that: 

\begin{equation}
\label{eq:1}
        ||f_{\theta_q}- f_\theta|| \leq L_\theta^\infty \delta_U,
\end{equation}

for some parameters $\theta$ and quantized parameters $\theta_q$ of a network $f$.  Here, the term $f_{\theta_q}$ is simply the perturbed velocity field when we are using parameters $\theta_q$ according to $f_{\theta + \Delta \theta}$. 

\begin{lemma} \label{lemma:1}
    Let $e_t = \hat{x}_t - x_t$, where $\hat{x}$ is the quantized flow driven by $\theta_q$, and $x$ is the flow driven by $\theta$. Then: 

    \begin{equation}
        \frac{d}{dt}||e_t|| \leq L_x||e_t|| + L_\theta^\infty \delta_U
    \end{equation}

    With the solution to the ODE being: 

    \begin{equation}
        ||e_t|| \leq \epsilon_U(t,b) := \frac{L_\theta^\infty \delta_U}{L_x}(e^{L_xt} - 1)
    \end{equation}

    Proof sketch: we set up two ODEs (for the quantized and non-quantized models) and use Assumption \ref{ass:1a} to obtain a scalar differential inequality. Using Gr\"{o}nwell's inequality, we come to the boundary cases. The full derivation has been left for the readers in the supplementary material.

\end{lemma}

\begin{lemma} \label{lemma:3}
    For every $t \in [0,T]$:

    \begin{equation}
        W_2(p_t, \hat{p}_t) \leq \epsilon_U(t,b) \implies \text{FID}(t) \leq L_\phi^2 \epsilon_U(t,b)^2
    \end{equation}
    
    This is a derivation by construction of Assumption \ref{ass:1e} and the definition of the Wasserstein-2 distance for two probability measures. 
\end{lemma}

Using the above lemmas and assumptions, we come to the following theorem. 
\begin{figure*}
    \centering
    \includegraphics[width=0.9\linewidth]{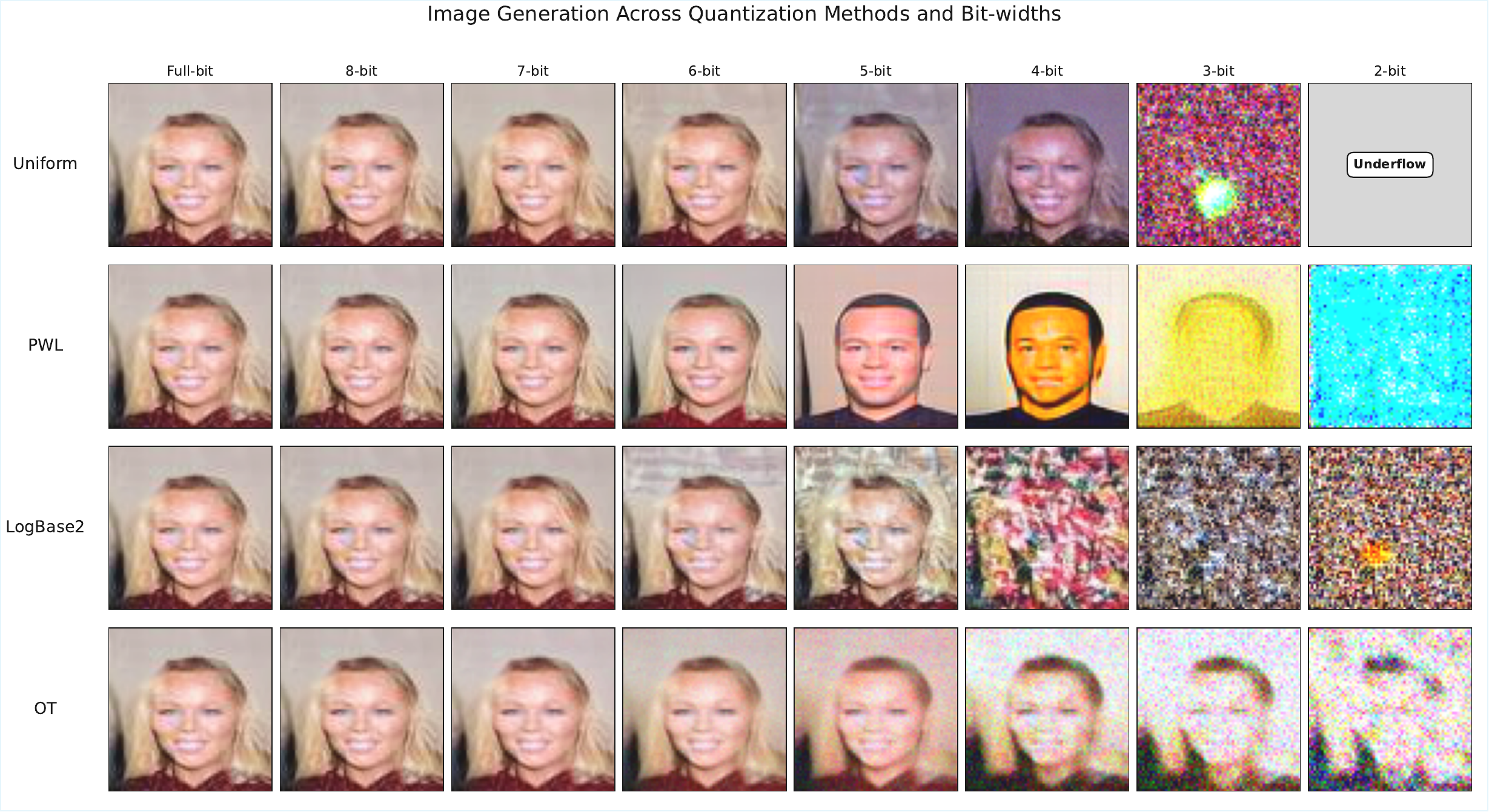}
    \caption{Comparison of OT-based quantization to other post-training quantization methods at various bit-widths, demonstrating the robustness and sampling quality for a model trained on the CelebA dataset.}
    \label{fig:celeba-demo}
\end{figure*}
\begin{theorem} \label{theorem:1}

    For a model that is uniformly quantized post-training, the FID is upper-bounded through
    \begin{equation}
        \text{FID}(T) \leq L_\phi^2\Bigg[ \frac{L_\theta^\infty}{L_x} (e^{L_x T } - 1) R  \Bigg]^2 2^{-2b}
    \end{equation}

    The proof of this emerges by substituting the values of $\delta_U$ and $\epsilon_U$ into Lemma \ref{lemma:3}. 
\end{theorem}

Beyond an upper bound, this allows us to see the correlation between the FID of a generative model and the bit-width, showing that $\text{FID}(T) \propto 2^{-2b}$. 

\subsection{Optimal-Transport Quantization}

This study adapts a non-uniform, equal-mass quantization approach deemed as Optimal Transport (OT)-based quantization. We interpret the full-precision trained weights of a given layer as samples drawn from an empirical distribution. Let $w_1, w_2, \cdots, w_N$ be the weights of a single layer (with $N$ parameters) for a flow matching model. We treat $P_w = \frac{1}{N}\sum_{i=1}^N \delta_{w_i}$ as the empirical distribution of these weights along a single dimension, resulting from the flattening of each weight matrix. Quantization seeks a set of $K$ representative values ${c_1, c_2, \dots, c_K}$ that are stored in a code-book, replacing the $N$ corresponding weights. This is equivalent to finding a distribution $Q = \sum_{j=1}^K p_j,\delta_{c_j}$ (with $\sum_j p_j=1$) that best approximates $P_w$. The error between the true weight distribution $P_w$ and $Q$ can be approximated using the 2-Wasserstein distance $W_2$. On $\mathbb{R}$, $W_2$ is defined as:
\begin{equation}
    W_2^2(P_w, Q) = \inf_\pi \int_{\mathbb{R} \times \mathbb{R}} (x-y)^2 d\pi(x,y)
\end{equation}
Where $x,y$ are drawn from the true and quantized weight distributions, respectively. $\pi$ is the joint (coupling) distribution with marginals $P_w$ and $Q$. Intuitively, $W_2^2(P_w,Q)$ is the minimum total squared transport cost to move the probability mass of $P_w$ to match $Q$. In our setting, since $P_w$ is discrete (each weight is a point mass $1/N$) and $Q$ has $K$ possible representations, this $W_2^2$ is exactly the average squared quantization error. Thus, finding the optimal quantization is equivalent to minimizing the 2-Wasserstein distance between the weight distribution and its quantized approximation. 

For distributions on $\mathbb{R}$, the optimal $W_2$ coupling is achieved by sorting the source and target distributions and pairing their quantile points. In other words, the mass at a given quantile of $P_w$ should be transported to the same quantile of $Q$. This result is a consequence of the Monge-Kantorovich theory in one dimension \cite{emami2024monge}, and implies that when approximating a 1D distribution using representative values, the representative values of $Q$ should correspond to contiguous segments of the sorted $P_w$ values.

In one dimension, the $W_2$-optimal quantization has an analytic solution. The classic Lloyd–Max quantization theory (which aligns with the above OT view) states that to minimize mean squared error, the real line should be partitioned into $K$ intervals each containing equal probability mass of $P_w$, and each quantization level should be the centroid (mean) of the weights in its interval. Let $F_w^{-1}(q)$ be the quantile function (inverse CDF) of the weight distribution. Then the optimal $K$-point approximation in the $W_2$ sense is achieved by choosing threshold quantiles $0 = q_0 < q_1 < \dots < q_K = 1$ such that $q_j - q_{j-1} = \frac{1}{K}$ for all $j$, with codebook values

\begin{equation}
    C_j = \mathbb{E} \Big[ w \mid F_w(w) \in [q_{j-1}, q_j]  \Big],
\end{equation}

for $j = 1, \cdots, K$. Equivalently, if $w_{(1)} \leq \cdots \leq w_{(N)}$ are the sorted weights, we partition this sorted list into $K$ groups of size $\approx N/K$ (as equal as possible), and set $c_j$ to the average of the $j^{\text{th}}$ group. Intuitively, each quantized value $c_j$ now carries an equal portion of the weight distributions' total probability. Since it is the mean of that segment, it is the best least-squares representative for the weights in that segment. This procedure automatically allocates finer resolution in high-density regions and coarser resolution in low-density regions. It is provably optimal for one-dimensional distributions under the quadratic cost: no other assignment of $N$ weights to $K$ levels can achieve lower average approximation error from the true weight distribution. The process is described in pseudo-code through Algorithm \ref{alg:ot_quant_no_clip}.

\begin{algorithm}
  \caption{OT\_Quantize$(W,b)$}
  \label{alg:ot_quant_no_clip}
  \begin{algorithmic}[1]
    \REQUIRE $W\in \mathbb{R}^{C\times d_1\times\cdots\times d_m}$, bit–width $b$
    \STATE $K \leftarrow 2^b$
    \FOR{$c=1$ \TO $C$}
      \STATE $\mathbf{v} \leftarrow \mathrm{Flatten}(W[c])$
      \STATE $\mathbf{v}_{\uparrow} \leftarrow \mathrm{sort}(\mathbf{v})$
      \STATE $\{B_k\}_{k=1}^{K} \leftarrow \mathrm{EqualMassSplit}(\mathbf{v}_{\uparrow},K)$
      \FOR{$k=1$ \TO $K$}
        \STATE $c_{c,k} \leftarrow \dfrac{1}{\lvert B_k\rvert}\sum_{w\in B_k} w$
      \ENDFOR
      \FOR{$i=1$ \TO $N$}
        \STATE $a_{c,i} \leftarrow \displaystyle\arg\min_{k}\lvert \mathbf{v}_i - c_{c,k}\rvert$
      \ENDFOR
    \ENDFOR
    \RETURN centroids $\{c_{c,k}\}_{c,k}$ and assignments $\{a_{c,i}\}_{c,i}$
  \end{algorithmic}
\end{algorithm}

\section{Theoretical Analysis}

Let the final, trained scalar weight density in a layer be $f_W$. If we partition the real line into $M = 2^b$ cells (denoted {$C_i$}) with equal probability, then we can say: 

\begin{equation}
    \int_{C_i} f_W(w) dw = \frac{1}{M}
\end{equation}

Here, the reconstruction levels are simply the cell or partition means. To get the mean-squared weight error, we use high-resolution quantization theory to get: 

\begin{equation}
    D_E = \frac{\alpha(f_W)^3}{12}2^{-2b}, \qquad \alpha(f_W) = \int_\mathbb{R} f_W(w)^{1/3} dw
\end{equation}

Although high-resolution quantization theory only applies for higher bit-widths, we use it here as a heuristic measure for comparisons to other quantization techniques. 

\begin{lemma}
    Using Assumption \ref{ass:1c}:

    \begin{equation}
        \mathbb{E}||f_{\theta_q}(x',t) - f_\theta(x,t)||^2 \leq (L_\theta^2)^2 pD_E
    \end{equation}

    With $p$ representing the number of independent noise sources hitting the network (one per weight) and $D_E$ being the average variance of each individual noise source (borrowed from the above definition). This is a consequence of $\mathbb{E}||\Delta \theta||_2^2$. 
\end{lemma}

\begin{lemma}
    Using a similar argument as in Lemma \ref{lemma:1}, we yield the following in means: 

    \begin{equation}
        \mathbb{E}||x_t - \hat{x}_t|| \leq \epsilon_E(t,b), \quad \epsilon_E(t,b) = \frac{L_\theta^2 \sqrt{pD_E}}{L_x} (e^{L_xt} - 1)
    \end{equation}

    The proof of this is identical to Lemma \ref{lemma:1} and is left in the supplementary items for readers. 
\end{lemma}

Adapting Lemma \ref{lemma:3} for an OT-based quantizer, we get $\text{FID}(t) \leq L_\phi^2\epsilon_E(t,b)^2$, which leads to:  

\begin{theorem} \label{theorem:2}
    \begin{equation}
        \text{FID}(T) \leq L_\phi^2 \Bigg[ \frac{L_\theta^2 \sqrt{p}}{L_x} (e^{L_xT} - 1)  \Bigg]^2 \frac{\alpha f_W(w)^3}{12} 2^{-2b}
    \end{equation}
\end{theorem}

\begin{figure*}
    \centering
    \begin{subfigure}[b]{0.95\textwidth}
        \centering
        \includegraphics[width=\textwidth]{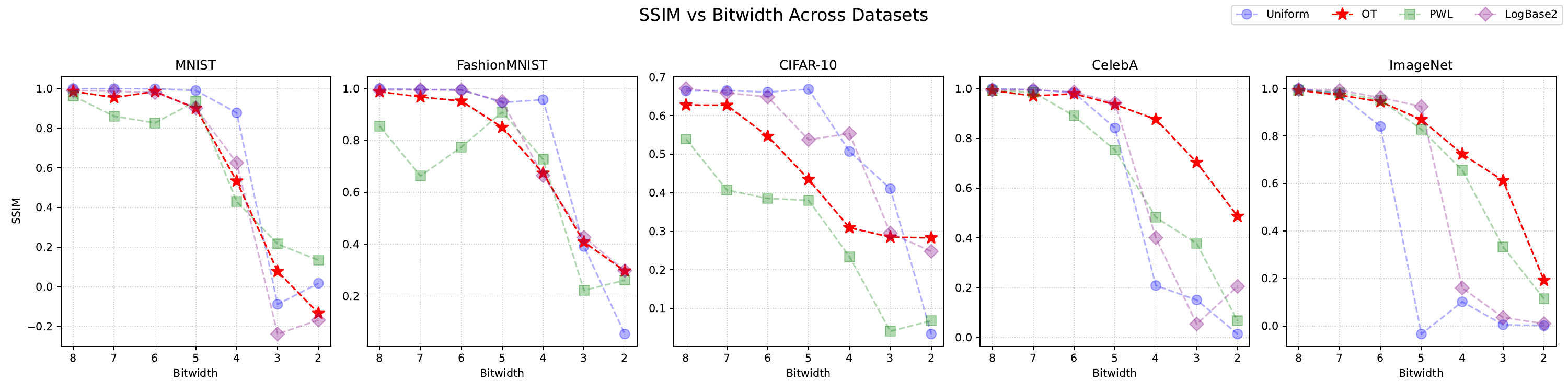}
        \caption{SSIM vs Bitwidth across datasets.}
        \label{fig:ssim}
    \end{subfigure}
    \vspace{0.7em}
    \begin{subfigure}[b]{0.95\textwidth}
        \centering
        \includegraphics[width=\textwidth]{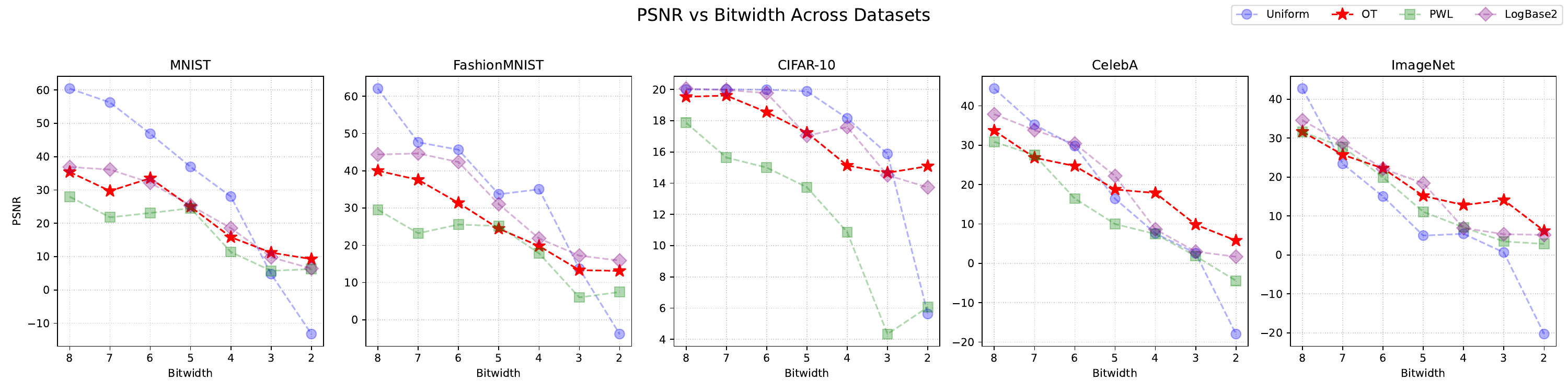}
        \caption{PSNR vs Bitwidth across datasets.}
        \label{fig:psnr}
    \end{subfigure}
    \caption{
        Quantitative evaluation of generative fidelity under quantization. (A) SSIM and (B) PSNR scores for each quantization scheme and bitwidth, evaluated across all benchmark datasets. OT quantization consistently outperforms alternatives, particularly at extreme low bit-widths.
    }
    \label{fig:quant_fidelity}
\end{figure*}
\subsection{Provable Advantages of OT-Quantization}

Collecting Theorems \eqref{theorem:1} and \eqref{theorem:2}, both post-training bounds have the same bit–width exponent but different front–constants:
\begin{equation}
  \text{FID}(T)
  \;\le\;
  \underbrace{C_U}_{\text{uniform}}
  \,2^{-2b}
  \;=\;
  \underbrace{C_E}_{\text{OT-based}}
  \,2^{-2b},
\end{equation}

We define
\begin{equation}
  \rho(b)\;:=\;\frac{C_E}{C_U}
  \;=\;
  \frac{\bigl(L_\theta^{2}\sqrt{p}\bigr)^{2}}{\bigl(L_\theta^{\infty}R\bigr)^{2}}
  \,\frac{\alpha(f_W)^3}{12}.
  \label{eq:rho}
\end{equation}
In practice, we expect \(L_\theta^{2}\sqrt{p}\!\approx\!L_\theta^{\infty}R\), so the ratio is
dominated by the histogram term
\(\alpha(f_W)^3/12\). For weights with Gaussian or Laplace tails,
        \(\alpha(f_W)^3\!\simeq\!0.3\,R^{2}\)
        $\;\Rightarrow\;$\(\rho\approx0.25\!-\!0.4\).
        OT-based quantization is therefore less than $2 \times$ tighter.

To determine why \(\alpha(f_W)^3\!\simeq\!0.3\,R^{2}\), assume the per–layer weights follow a Gaussian distribution with
\[
f_W(w)=\frac{1}{\sqrt{2\pi}\,\sigma}\,
       \exp\!\Bigl(-\tfrac{w^{2}}{2\sigma^{2}}\Bigr).
\]
The OT constant is defined by
\begin{equation}
\begin{aligned}
\alpha(f_W)
  &= \int_{-\infty}^{\infty} f_W(w)^{1/3}\,dw \\
  &= (\sqrt{2\pi}\,\sigma)^{-1/3}
     \int_{-\infty}^{\infty}
     \exp\left(-\frac{w^2}{6\sigma^2}\right) dw \\
  &= \frac{\sqrt{6\pi}}{(2\pi)^{1/6}}\, \sigma^{2/3} = 32.8\,\sigma^{2/3}.
\end{aligned}
\end{equation}

Uniform PTQ picks a symmetric range \(R\) that covers essentially all weights. A common choice is the \(k\sigma\) rule with \(k\!\in[8,10]\). Substituting \(R=k\sigma\) gives
$\frac{\alpha(f_W)^{3}}{R^{2}} =\frac{32.8}{k^{2}}, k=10 \Rightarrow\;\mathbf{0.33}.$
Thus, for the \(\;k\!\simeq\! 10  \sigma\) clipping level,
\[
\boxed{\;
\alpha(f_W)^{3}\;\approx\;0.3\,R^{2}}\quad(\text{Gaussian}).
\]

For the two-sided Laplace probability density function,   
\(f_W(w)=\tfrac{1}{2\beta}\,e^{-|w|/\beta}\)
(with standard deviation \(\sigma=\sqrt{2}\,\beta\)):
\[
\alpha(f_W)^{3}=108\,\beta^{2}=54\,\sigma^{2}.
\]
Using the same \(k\sigma\) range,
\[
\frac{\alpha(f_W)^{3}}{R_{\text{PTQ}}^{2}}
   =\frac{54}{k^{2}}
   \;\xrightarrow{k=10}\;0.54,
\]
Therefore, sub-Gaussian (Gaussian / Laplace) weight histograms and
layer-wise PTQ ranges that cover \(8{-}10\,\sigma\) would result in
$\alpha(f_W)^{3}\;\;\approx\;(0.3\!-\!0.5)\,R^{2}$ explaining why the equal-mass constant \(C_E\propto\alpha(f_W)^{3}\) is typically \(\sim2\times\) smaller than the uniform constant \(C_U\propto R^{2}\).

\paragraph{Intuition}
Uniform PTQ must set \(R\) large enough for the single
largest weight, which that inflates every bin width \(\Delta\) and hence every per-weight error \(\delta_U\). Equal-mass quantization, on the other hand, allocates only \(\frac{1}{M}=2^{-b}\) probability to the tail cell, so the tail contributes negligibly to the average MSE. Other quantization methods follow similar proportionalities to $2^{-2b}$ with differing front-constants. 

\section{Empirical Findings}
\begin{figure*}
    \centering
    \includegraphics[width=\linewidth]{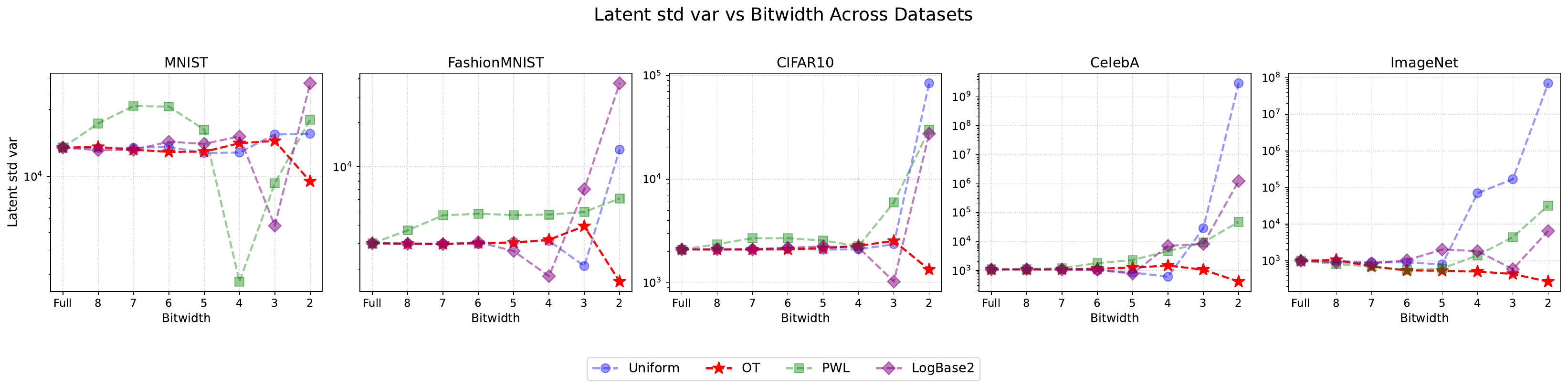}
    \caption{Latent variance standard deviation versus bitwidth for each quantization method and dataset. OT quantization maintains stable latent structure across all bitwidths, while alternative methods become increasingly unstable at low bits, especially for complex datasets.}
    \label{fig:latent_std}
\end{figure*}
For empirical evaluation, we used the standard Flow Matching implementation from Meta AI and adopted the recommended hyperparameters as specified in \cite{lipman2024flowmatchingguidecode}. Experiments were conducted on five benchmark datasets: MNIST \cite{lecun2002gradient}, FashionMNIST \cite{xiao2017fashion}, CIFAR10 \cite{cifar}, ImageNet \cite{imagenet}, and CelebA \cite{celeba}, spanning a wide range of dataset complexities in terms of the number of classes, degree of visual diversity, and intra-class variation. Models were trained on a single A100 GPU, with specific training statistics and details given within the linked GitHub repository.

\paragraph{Generation Quality} To assess the impact of quantization with respect to output quality, we compared images generated by full-precision and quantized FM models using two established metrics: Peak Signal-to-Noise Ratio (PSNR) for pixel-level accuracy and Structural Similarity Index Measure (SSIM) for perceptual quality. For each method and bit-width, average PSNR and SSIM scores were computed against the full-precision reference outputs as illustrated in Figure \ref{fig:celeba-demo}, with the reference set corresponding to the evaluation split of each dataset.

As depicted in Figure \ref{fig:quant_fidelity}, OT-quantization seems to demonstrate consistently higher SSIM (\ref{fig:ssim}) and PSNR (\ref{fig:psnr}) scores than conventional baselines (Uniform, PWL, LogBase2), with its advantages most evident at lower bit-widths. For more complex datasets—characterized by greater visual diversity and class cardinality, such as CelebA and ImageNet—alternative methods show accelerated degradation in perceptual and signal fidelity below the 5-bit threshold. OT quantization, by contrast, maintains relatively stable performance and avoids abrupt fidelity loss even at 2–3 bits, suggesting a greater resilience to quantization-induced artifacts. Although the absolute improvements are moderate, the trend supports the theoretical expectation that OT-based quantization can better preserve generative fidelity under aggressive compression. These empirical results reinforce the practical relevance of OT quantization for deploying flow matching generative models in resource-constrained environments, where even marginal improvements in robustness can have significant operational value.

\paragraph{Latent Space Stability} Assessing disentanglement in large scale real world datasets such as CelebA and ImageNet directly remains an open challenge \cite{carbonneau2022measuring}. Classical metrics like MIG \cite{chen2018isolating}, SAP \cite{kumar2017variational}, and $\beta$-VAE require access to explicit generative factors, as well as the assumption of low interfactor correlation. These conditions are rarely met outside of synthetic or heavily curated benchmarks. In our setting, where the underlying factors are unknown or exhibit significant dependencies, these disentanglement metrics become impractical or uninterpretable.

Instead, we focus on the stability of the learned latent space under quantization. As shown in Figure~\ref{fig:latent_std}, OT quantization uniquely maintains both the mean and standard deviation of latent variances across all tested datasets and bitwidths. By contrast, Uniform and LogBase2 quantization lead to dramatic increases in latent variance dispersion at lower bitwidths, reflecting latent space destabilization and the amplification of quantization noise. This effect is especially pronounced on complex datasets, where standard quantization schemes exhibit variance explosions that do not occur with OT quantization.

These findings provide a cohesive picture: OT based quantization not only delivers modest but consistent improvements in perceptual and signal fidelity, but also uniquely preserves the internal structure of the generative process even under aggressive bitwidth reduction. This dual stability, across both output and latent representations, explains the empirical robustness of OT quantized models and underscores their suitability for deployment on edge devices or in distributed inference scenarios, where quantization budgets are stringent.

Taken together, these results support the adoption of OT quantization as a principled default for compressing flow matching generative models, particularly in domains where performance reliability under extreme compression is essential. 

\section{Future Work and Conclusion}
This study establishes that optimal transport quantization not only preserves output fidelity but also maintains latent space stability in flow matching generative models, even under extreme compression. While these results strongly motivate the use of OT quantization for resource-constrained deployment, several important research directions remain.

Future work should investigate how the observed latent space regularity induced by OT quantization impacts downstream tasks, including sample diversity, mode coverage, and robustness to out of distribution data. Another promising direction is to examine the interplay between quantization and self-supervised or semi-supervised fine tuning, which may further improve resilience to precision loss. It is also important to assess the generalizability of OT quantization to other generative architectures, such as diffusion transformers and latent variable models, and to evaluate its impact on practical deployment metrics such as inference speed and energy efficiency in edge scenarios.

In addition, further studies should systematically analyze codebook utilization and quantization level efficiency under OT quantization, ensuring effective representation within the reduced precision parameter space. Rigorous evaluation of hardware compatibility, including support for low bit-width arithmetic and efficient on-chip quantization routines, is crucial for bridging the gap between algorithmic advances and real world deployment. Direct measurement of power consumption, latency, and memory bandwidth on representative edge devices will provide a more realistic assessment of operational benefits and constraints. Finally, expanding the theoretical framework for OT quantization to characterize generalization properties and its behavior under diverse data distributions would offer deeper insight and guide future algorithmic development.

\bibliography{biblio.bib}
\newpage

\section{Appendix}
In this supplementary material, we elaborate on the proofs needed for the main lemmas, theorems and derivations used for the submitted paper ``Low-Bit, High-Fidelity: Optimal Transport Quantization for Flow Matching.'' Although the sketches for important proofs have been outlined in the main manuscript, this document is intended to use for complete derivations for interested researchers and reviewers. Additionally, we provide additional generated samples for other datasets for illustration purposes and alignment with the numerical findings.  

\subsection{Definitions}
\begin{definition}
    For a weight matrix $W$ in a network, uniform PTQ produces a symmetric range characterized by $[-R, R]$ covering all weights in the layer. On a more granular scale, this can also be defined per-channel for some $w^{(n)} \in W$ for $W\in \mathbb{R}^{n \times m}$. With $b$ bits, the step-size $\Delta$ is defined as: 

    \begin{equation}
        \Delta = \frac{2R}{2^b}
    \end{equation}
\end{definition}

\begin{definition} \label{def:worst-case}
    We define the worst-case uniform quantization error per individual weight $w \in W$ as $\delta_U$.

    \begin{equation}
        \delta_U \leq \frac{\Delta}{2} = \frac{R}{2^{b-1}}
    \end{equation}

    This gives the maximum perturbation that a weight $w$ can have from the ``true'' value. This directly follows as a result of the reconstruction given a quantization bin.
\end{definition}

\subsection{Proofs}
We first treat worst-case (uniform) error bounds, and then average (OT) error bounds. Under Assumption \ref{ass:1b}, it follows that: 

\begin{equation}
\label{eq:1}
        ||f_{\theta_q}- f_\theta|| \leq L_\theta^\infty \delta_U,
\end{equation}

\begin{proof}
    Let $\theta_q$ be the quantized parameters for the network $f$. Here, the term $f_{\theta_q}$ is simply the perturbed velocity field when we are using parameters $\theta_q$ according to $f_{\theta + \Delta \theta}$. 
        
    This thus becomes a direct consequence of the assumption. 
\end{proof}

\begin{lemma} \label{lemma:2}
    Let $e_t = \hat{x}_t - x_t$, where $\hat{x}$ is the quantized flow driven by $\theta_q$, and $x$ is the flow driven by $\theta$. Then: 

    \begin{equation}
        \frac{d}{dt}||e_t|| \leq L_x||e_t|| + L_\theta^\infty \delta_U
    \end{equation}

    With the solution to the ODE being: 

    \begin{equation}
        ||e_t|| \leq \epsilon_U(t,b) := \frac{L_\theta^\infty \delta_U}{L_x}(e^{L_xt} - 1)
    \end{equation}

    \begin{proof}

        We set up the two ODEs (for the quantized and non-quantized models) as follows: 

        \begin{equation*}
            \begin{split}
                \frac{d}{dt} x_t = f_\theta(x,t), \quad x_0 \sim p_0\\
                \frac{d}{dt} \hat{x}_t =  f_{\theta_q}(x,t), \quad \hat{x}_0 = x_0
            \end{split}
        \end{equation*}

        Since both trajectories start from the same $x_0$, then $e_0 = 0$ by definition. Then: 

        \begin{equation*}
            \begin{split}
                \frac{d}{dt}e_t = \frac{d}{dt}(\hat{x}_t - x_t) = \frac{d}{dt} \hat{x}_t - \frac{d}{dt} x_t = f_{\theta_q}(x,t) - f_\theta (x, t) \\
                \frac{d}{dt}e_t = f_{\theta_q}(x,t) - f_{\theta_q}(x,t) + f_{\theta_q}(x,t) - f_\theta (x, t) \\
            \end{split}
        \end{equation*}

        We split the term by adding and subtracting $f_{\theta_q}(x,t)$. Now, under Assumption \ref{ass:1a}, 

        \begin{equation*}
            || f_{\theta_q}(x,t) - f_{\theta_q}(x,t)|| \leq L_x||\hat{x}_t - x_t|| = L_x||e_t||
        \end{equation*}

        Similarly, using \eqref{eq:1}:
        \begin{equation*}
            ||f_{\theta_q}- f_\theta|| \leq L_\theta^\infty \delta_U
        \end{equation*}

        By taking the norm, we can obtain a scalar differential inequality (triangular inequality): 

        \begin{equation*}
            \frac{d}{dt}||e_t|| \leq L_x||e_t|| + L_\theta^\infty \delta_U \quad \forall t\geq 0, \quad ||e_0|| = 0
        \end{equation*}

        This is a first order linear differential inequality in the scalar $y(t) = ||e_t||$. We solve the ODE $y'(t) = L_xy(t) + L_\theta^\infty \delta_U$ with $y(0) = 0$ (by integrating factor): 

        \begin{equation*}
            y(t) = \frac{L_\theta^\infty \delta_U}{L_x}(e^{L_x t} - 1) := \epsilon_U(t,b)
        \end{equation*}

        Since $|| f_{\theta_q}(x,t) - f_{\theta_q}(x,t)|| \leq L_x||e_t||$ is an upper inequality, Gronwall's lemma tells us that the actual $y(t) = ||e_t||$ cannot exceed this solution. Therefore, 

        \begin{equation*}
            ||e_t|| \leq \epsilon_U(t,b)
        \end{equation*}

        We check for boundary cases: 
        \begin{itemize}
            \item If $L_x = 0$, then the upper bound reduces smoothly to $L_\theta^\infty \delta_U (t)$ by taking the limit as $L_x \rightarrow 0$. 
            \item If $\delta_U = 0$ (i.e., there is no quantization), then $\epsilon_U(t,b) = 0$ and the flows coincide, as expected. 
        \end{itemize}
    \end{proof}

\end{lemma}

\begin{lemma} \label{lemma:3}
    For every $t \in [0,T]$:

    \begin{equation*}
        W_2(p_t, \hat{p}_t) \leq \epsilon_U(t,b) \implies \text{FID}(t) \leq L_\phi^2 \epsilon_U(t,b)^2
    \end{equation*}

    In plain language, we say that the trajectory error bound obtained through Lemma \ref{lemma:2} allows us to obtain the Wasserstein-2 bound, which is then used as a proxy for the FID. 

    \begin{proof}
        By definition, the Wasserstein-2 ($W_2$) is given for two probability measures $\mu, \nu$ on $\mathbb{R}^d$ by: 

        \begin{equation*}
            W_2^2(\mu, \nu) = \inf_{\gamma \in \prod(\mu, \nu)} \int_{\mathbb{R}^d \times \mathbb{R}^d} ||x-y||^2 d\gamma(x,y) 
        \end{equation*}

        Where $\prod(\mu, \nu)$ is the set of couplings whose marginals are $\mu$ and $\nu$, with $x,y$ sampled from $\mu$ and $\nu$, respectively. 

        Since we already have a canonical joint construction of $(x_t, \hat{x}_t)$, we can construct a specific coupling by denoting $\gamma_t$ as the law of the pair $(x_t, \hat{x}_t)$. By construction,  $\gamma_t \in \prod(p_t, \hat{p}_t)$ for $p_t$ and $\hat{p}_t$ being the laws of $x_t, \hat{x}_t$, respectively.

        Since the infimum is less or equal to the value at any admissible $\gamma$, then: 

        \begin{equation*}
            W_2^2(p_t, \hat{p}_t) \leq \int ||x-\hat{x}||^2 d\gamma(x, \hat{x}) = \mathbb{E}||x_t - \hat{x}_t||^2
        \end{equation*}

        We use Lemma \ref{lemma:2} to get the deterministic bound: $||x_t - \hat{x}_t|| \leq \epsilon_U(t,b)$ path-wise. If we square both sides, we obtain: 

        \begin{equation*}
            \mathbb{E}||x_t - \hat{x}_t||^2 \leq \epsilon_U(t,b)^2
        \end{equation*}

        We combine both facts to thus get: 

        \begin{equation*}
            W_2^2(p_t, \hat{p}_t) \leq \epsilon_U(t,b)^2
        \end{equation*}

        This completes the first part of the proof. In the second part, we will relate the FID to the $W_2$ bound. 

        Define $\mu_t = \phi(p_t)$ and $\nu_t = \phi(\hat{p}_t)$ as the push-forwards of the two laws by the Inception-v3 feature extractor, $\phi$. By Assumption \ref{ass:1d}, $\phi$ is $L_\phi$-Lipschitz, meaning that: 

        \begin{equation*}
            ||\phi(x) - \phi(x')|| \leq L_\phi ||x-x'|| \quad \forall x, x'
        \end{equation*}

        A standard property of optimal transport then gives: 

        \begin{equation*}
            W_2(\mu_t, \hat{\mu}_t) \leq L_\phi W_2(p_t, \hat{p}_t)
        \end{equation*}

        For any $\gamma_t \in \prod(p_t, \hat{p}_t)$, if we push this through $\phi$ to obtain a coupling for $\mu_t, \hat{\mu}_t$, the Lipschitz condition scales the cost by at most $L_\phi^2$. By inserting the bound obtained above: 

        \begin{equation*}
            W_2(\mu_t, \hat{\mu}_t) \leq L_\phi \epsilon_U(t, b)
        \end{equation*}

        Finally, using $W_2$ as a proxy for the FID by Assumption \ref{ass:1e}, we take the embedding distributions as approximately Gaussian. For two Gaussian distributions $\mathcal{N}(m, \Sigma)$ and $\mathcal{N}(m', \Sigma')$, the FID is calculated as: 

        \begin{align*}
            \text{FID} = \\
            ||m - m'||^2 +  \text{Tr}(\Sigma + \Sigma' - 2[\Sigma^{1/2} \Sigma' \Sigma^{1/2}]^{1/2})
            = W_2^2 (\mu_t, \hat{\mu}_t) \\
        \end{align*}

        Using this assumption, it directly follows that: 

        \begin{equation*}
            \text{FID}(t) \leq L_\phi^2\epsilon_U(t,b)^2
        \end{equation*}

        This concludes the proof.
    \end{proof}
\end{lemma}

\begin{theorem} \label{theorem:1}
    \begin{equation}
        \text{FID}(T) \leq L_\phi^2\Bigg[ \frac{L_\theta^\infty}{L_x} (e^{L_x T } - 1) R  \Bigg]^2 2^{-2b}
    \end{equation}

    \begin{proof}
        We plug in the values of $\delta_U$ and $\epsilon_U$ into the previous lemma. 

        Let $t=T$. Then, using \ref{lemma:3}:

        \begin{equation*}
            \text{FID}(T) \leq L_\phi^2\epsilon_U(T,b)^2
        \end{equation*}
        
        Since $\epsilon_U(t,b) := \frac{L_\theta^\infty \delta_U}{L_x}(e^{L_xt} - 1)$, then: 

        \begin{equation*}
            \text{FID}(T) \leq L^2_\phi\Big(\frac{L_\theta^\infty}{L_x} \delta_U (e^{L_xT}-1)\Big)
        \end{equation*}

        We insert the worst-case sensitivity / quantization error: 
        \begin{equation*}
            \delta_U \leq \frac{R}{2^{b-1}}
        \end{equation*}

        Bringing us to: 

        \begin{equation*}
            \text{FID}(T) \leq L_\phi^2\Bigg[ \frac{L_\theta^\infty}{L_x} (e^{L_x T } - 1) R  \Bigg]^2 2^{-2b}
        \end{equation*}

        Although an explicit factor of 4 would be added here through the collection of the powers of $2$, it is disregarded here as we absorb it into $R$ to provide a more ``relaxed'' bound. 

        This concludes the proof. 
        
    \end{proof}
\end{theorem}

Let the final, trained scalar weight density in a layer be $f_W$. If we partition the real line into $M = 2^b$ cells (denoted {$C_i$}) with equal probability, then we can say: 

\begin{equation}
    \int_{C_i} f_W(w) dw = \frac{1}{M}
\end{equation}

Here, the reconstruction levels are simply the cell or partition means. To get the mean-squared weight error, we use high-resolution quantization theory (Bennet's integral) to get: 

\begin{equation}
    D_E = \frac{\alpha(f_W)^3}{12}2^{-2b}, \qquad \alpha(f_W) = \int_\mathbb{R} f_W(w)^{1/3} dw
\end{equation}

\begin{lemma}
\label{lemma:lee}
    Using Assumption \ref{ass:1c}:

    \begin{equation}
        \mathbb{E}||f_{\theta_q}(x,t) - f_\theta(x,t)||^2 \leq (L_\theta^2)^2 pD_E
    \end{equation}

    With $p$ representing the number of independent noise sources hitting the network (one per weight) and $D_E$ being the average variance of each individual noise source (borrowed from the above definition). This is a consequence of $\mathbb{E}||\Delta \theta||_2^2$. 

    \begin{proof}
        Assumption \ref{ass:1c} states: 

        \begin{equation*}
            ||f_{\theta + \Delta \theta}- f_\theta|| \leq L_\theta^2||\Delta \theta||_2
        \end{equation*}

        By squaring both sides (using the well-known fact that if $a\leq b$ then $a^2 \leq b^2$): 

        \begin{equation*}
            ||f_{\theta_q}- f_\theta||^2 \leq (L_\theta^2)^2||\Delta \theta||_2^2
        \end{equation*}

        By taking the expectation on both sides, we obtain: 

        \begin{equation*}
            \mathbb{E}||f_{\theta_q}- f_\theta||^2 \leq (L_\theta^2)^2 \mathbb{E}||\Delta \theta||_2^2
        \end{equation*}

        Each weight perturbation $\Delta \theta_i$ is an independent, zero-mean random variable with variance $D_E$. Hence: 

        \begin{equation*}
            \mathbb{E}||\Delta \theta||_2^2=\mathbb{E}\Big[\sum_{i=1}^p\Delta\theta_i^2 \Big] = \sum_{i=1}^p \mathbb{E}\Delta\theta_i^2 = pD_E
        \end{equation*}

        By combining the above, we arrive at: 

        \begin{equation*}
            \mathbb{E}||f_{\theta_q}(x,t) - f_\theta(x,t)||^2 \leq (L_\theta^2)^2 pD_E
        \end{equation*}
    \end{proof}
\end{lemma}

\begin{lemma}
    Using a similar argument as in Lemma \ref{lemma:2}, we yield the following in means: 

    \begin{equation}
        \mathbb{E}||x_t - \hat{x}_t|| \leq \epsilon_E(t,b), 
        \quad \epsilon_E(t,b) = \frac{L_\theta^2 \sqrt{pD_E}}{L_x} (e^{L_xt} - 1) 
    \end{equation}

    \begin{proof}
        Consider the two flows:
        \begin{equation*}
            \frac{d}{dt}x_t = f_\theta(x_t, t), \quad x_0 \sim p_0,
        \end{equation*}
        \begin{equation*}
            \frac{d}{dt}\hat{x}_t = f_{\theta_q}(\hat{x}_t, t), \quad \hat{x}_0 = x_0,
        \end{equation*}
        and define $e_t = \hat{x}_t - x_t$, so that $e_0 = 0$.

        Differentiating $e_t$ gives:
        \begin{equation*}
            \dot{e}_t = f_{\theta_q}(\hat{x}_t, t) - f_\theta(x_t, t).
        \end{equation*}
        Add and subtract $f_{\theta_q}(x_t, t)$ to split the difference:
        \begin{equation*}
            \dot{e}_t =
            \underbrace{f_{\theta_q}(\hat{x}_t, t) - f_{\theta_q}(x_t, t)}_{\text{state term}}
            + \underbrace{f_{\theta_q}(x_t, t) - f_\theta(x_t, t)}_{\text{parameter term}}.
        \end{equation*}

        Taking norms, expectations, and applying the triangle inequality:
        \begin{equation*}
            \frac{d}{dt} \mathbb{E}\|e_t\|
            \leq
            \mathbb{E}\|f_{\theta_q}(\hat{x}_t, t) - f_{\theta_q}(x_t, t)\|
            +
            \mathbb{E}\|f_{\theta_q}(x_t, t) - f_\theta(x_t, t)\|.
        \end{equation*}

        \textbf{State term:} By Assumption~\ref{ass:1a}, 
        \[
            \|f_{\theta_q}(\hat{x}, t) - f_{\theta_q}(x, t)\| \leq L_x \|\hat{x} - x\| = L_x \|e_t\|,
        \]
        and hence
        \[
            \mathbb{E}\|f_{\theta_q}(\hat{x}_t, t) - f_{\theta_q}(x_t, t)\| \leq L_x\,\mathbb{E}\|e_t\|.
        \]

        \textbf{Parameter term:} From Lemma~\ref{lemma:lee}, 
        \[
            \mathbb{E}\|f_{\theta_q} - f_\theta\|^2 \leq (L_\theta^2)^2 p D_E.
        \]
        By Jensen's inequality, 
        \[
            \mathbb{E}\|f_{\theta_q} - f_\theta\| \leq L_\theta^2 \sqrt{p D_E}.
        \]

        Combining both terms, we obtain the scalar differential inequality:
        \begin{equation*}
            \frac{d}{dt} \mathbb{E}\|e_t\| \leq L_x\,\mathbb{E}\|e_t\| + L_\theta^2 \sqrt{p D_E}, \quad \mathbb{E}\|e_0\| = 0.
        \end{equation*}

        Let $y(t) = \mathbb{E}\|e_t\|$. The corresponding equality $y'(t) = L_x y(t) + C$, with $C = L_\theta^2 \sqrt{p D_E}$ and $y(0) = 0$, has the solution:
        \begin{equation*}
            y(t) = \frac{C}{L_x} \left(e^{L_x t} - 1\right).
        \end{equation*}
        By Grönwall's lemma, the inequality solution cannot exceed this value. Therefore:
        \begin{equation*}
            \mathbb{E}\|e_t\| \leq \frac{L_\theta^2 \sqrt{p D_E}}{L_x} \left(e^{L_x t} - 1\right) =: \epsilon_E(t,b).
        \end{equation*}
        This concludes the proof.
    \end{proof}
\end{lemma}

\begin{lemma}
\label{leeee}
    With non-uniform, equal-mass quantization, the FID is bounded by: 

    \begin{equation}
        \text{FID}(t) \leq L_\phi^2\epsilon_E(t,b)^2
    \end{equation}
\end{lemma}

\begin{theorem} \label{theorem:2}
    \begin{equation}
        \text{FID}(T) \leq L_\phi^2 \Bigg[ \frac{L_\theta^2 \sqrt{p}}{L_x} (e^{L_xT} - 1)  \Bigg]^2 \frac{\alpha (f_W(w))^3}{12} 2^{-2b}
    \end{equation}
\end{theorem}

Derivations for \ref{leeee} and \ref{theorem:2} are exactly the same as \ref{theorem:1} and have therefore been omitted for brevity.

Both post-training bounds have the same bit–width exponent but different front–constants:
\begin{equation}
  \text{FID}(T)
  \;\le\;
  \underbrace{C_U}_{\text{uniform}}
  \,2^{-2b}
  \;=\;
  \underbrace{C_E}_{\text{equal-mass}}
  \,2^{-2b},
\end{equation}
with
\[
  C_U=L_\phi^2\!\left[\frac{L_\theta^{\infty}}{L_x}(e^{L_xT}-1)R\right]^2,
\]
and 
\[
C_E=L_\phi^2\!\left[\frac{L_\theta^{2}\sqrt{p}}{L_x}(e^{L_xT}-1)\right]^2 \frac{\alpha(f_W)^3}{12}.
\]
\section{Two Immediate Corollaries}

\begin{corollary}[Bit budget]
  Fix a target FID gap $\Delta_{\max}$.  
  Any $b$ satisfying
  \[
    2^{-2b}\le\frac{\Delta_{\max}}{C_U}
    \quad(\text{uniform}),\qquad
    2^{-2b}\le\frac{\Delta_{\max}}{C_E}
    \quad(\text{OT})
  \]
  is guaranteed not to violate the budget.  
  Because \(C_E\approx\!\rho C_U\) with \(\rho<1\), equal-mass
  typically admits either \textit{two extra bits of headroom} or, phrased
  differently, the \emph{same} head-room at less than half the bit-width.
\end{corollary}

\begin{corollary}[Target FID vs.\ bit-width]
  Rearranging the bounds gives
  \begin{align*}
    b \;\ge\;
    \frac12\log_2\!\left(\frac{C_U}{\text{FID}_{\text{goal}}}\right)
    &\quad\text{(uniform)}, \\
    b \;\ge\;
    \frac12\log_2\!\left(\frac{C_E}{\text{FID}_{\text{goal}}}\right)
    &\quad\text{(OT)}.
  \end{align*}
\end{corollary}

\subsection{Generated Samples}
\begin{figure}[h]
    \centering
\includegraphics[width=0.9\linewidth]{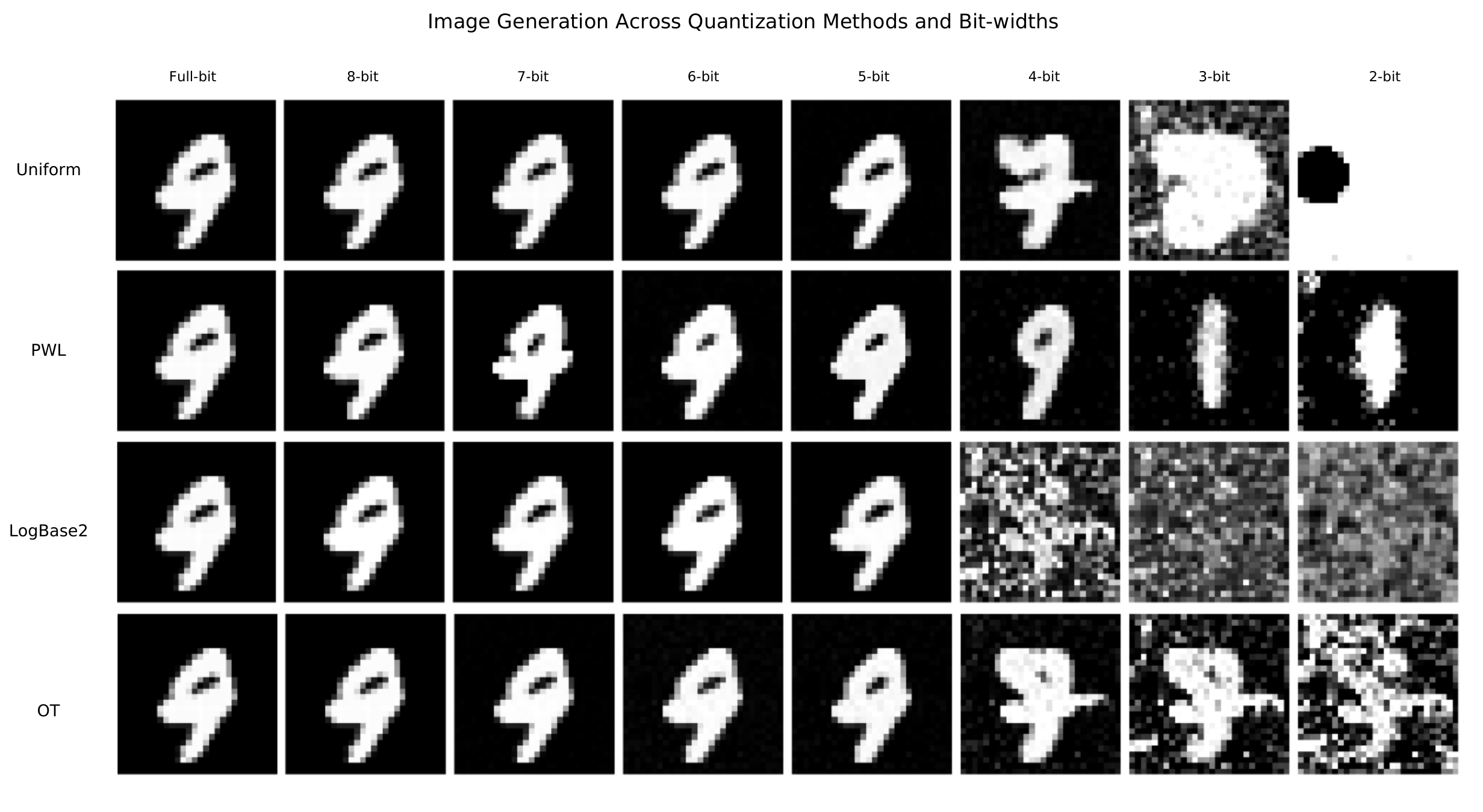}
    \caption{Generated samples for MNIST.}
    \label{fig:1}
\end{figure}

\begin{figure}[h]
    \centering
\includegraphics[width=0.9\linewidth]{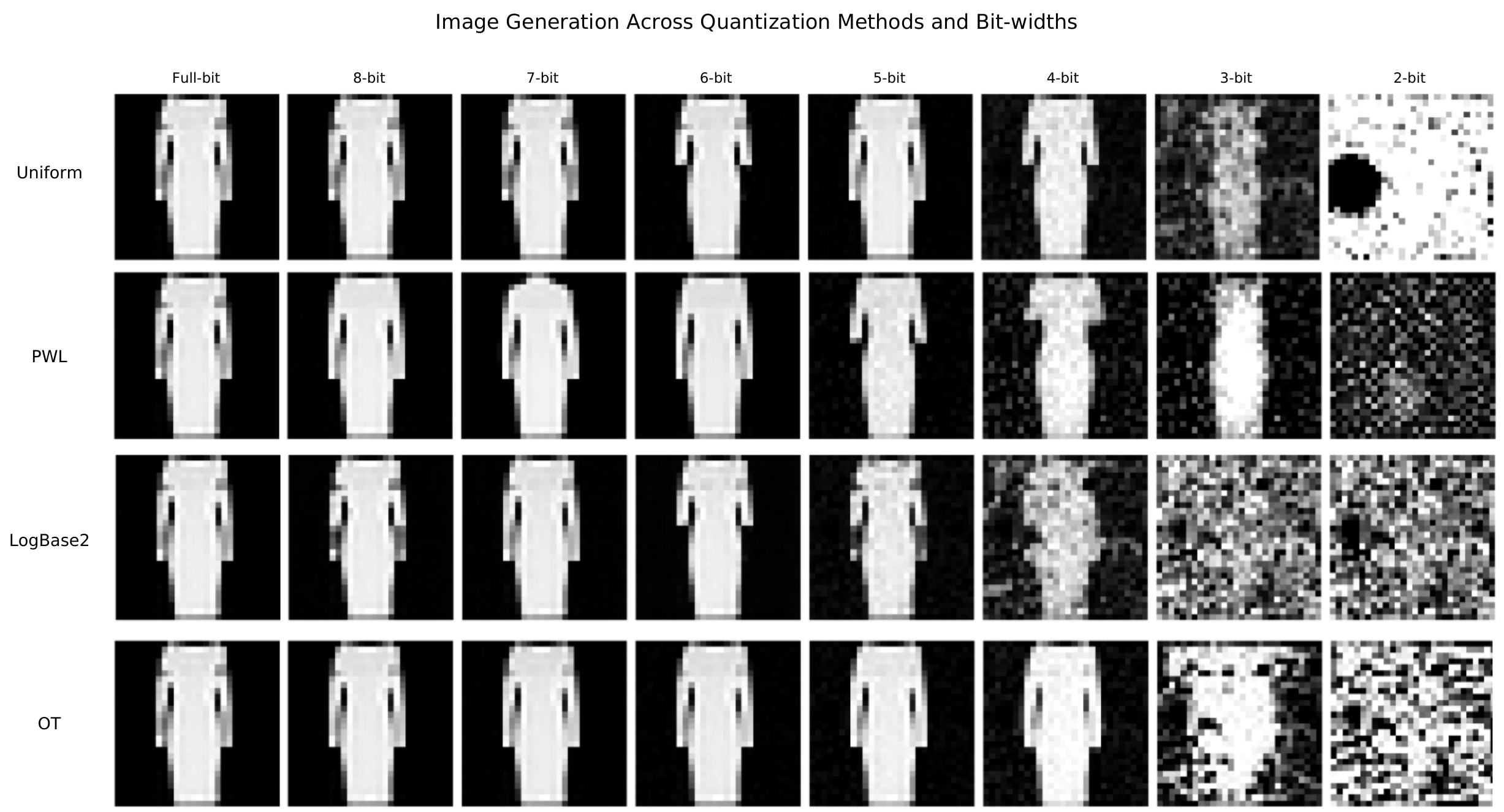}
    \caption{Generated samples for FashionMNIST.}
    \label{fig:2}
\end{figure}

\begin{figure}[h]
    \centering
\includegraphics[width=0.9\linewidth]{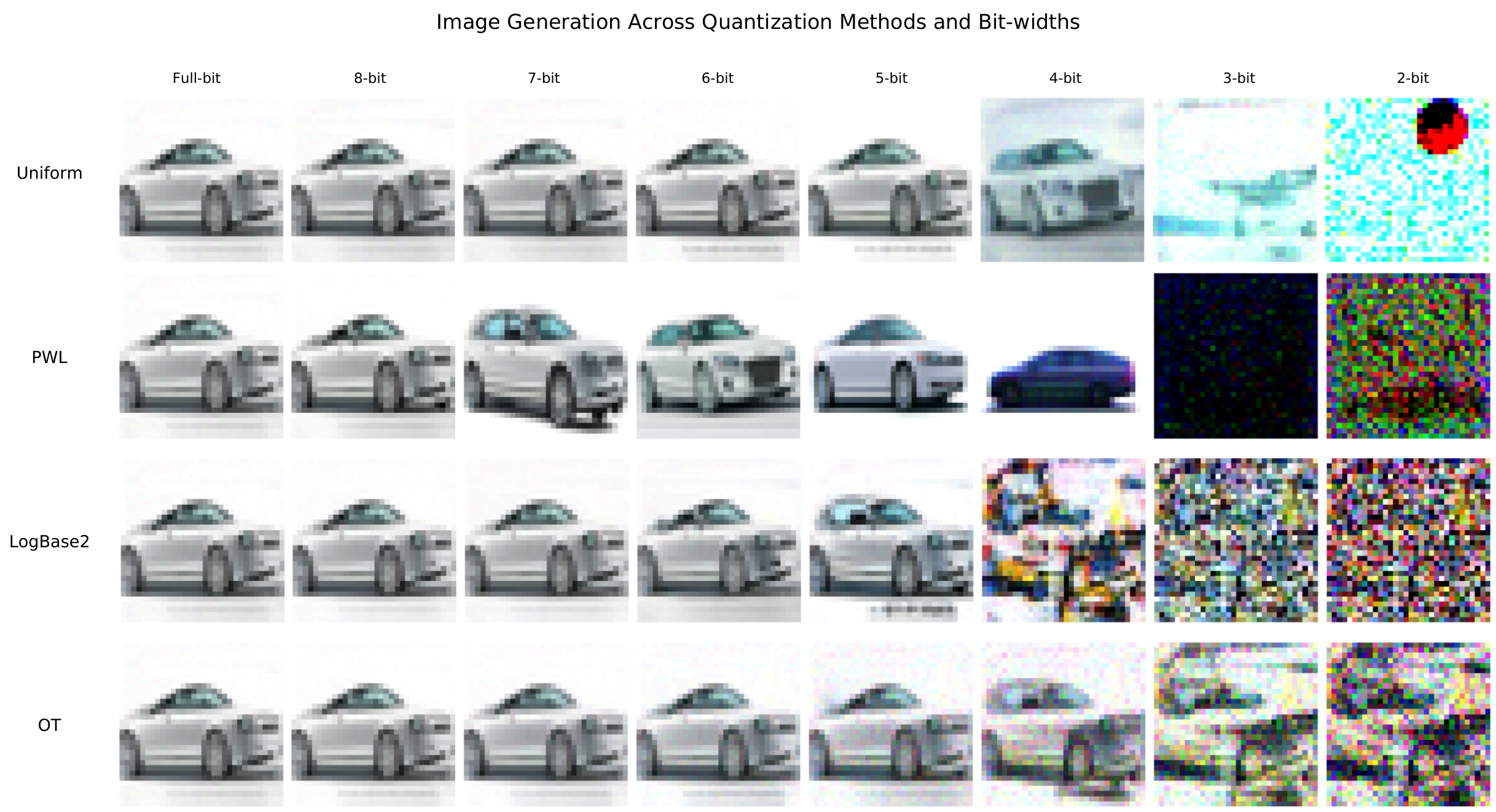}
    \caption{Generated samples for CIFAR 10.}
    \label{fig:3}
\end{figure}

\begin{figure}[h]
    \centering
\includegraphics[width=0.9\linewidth]{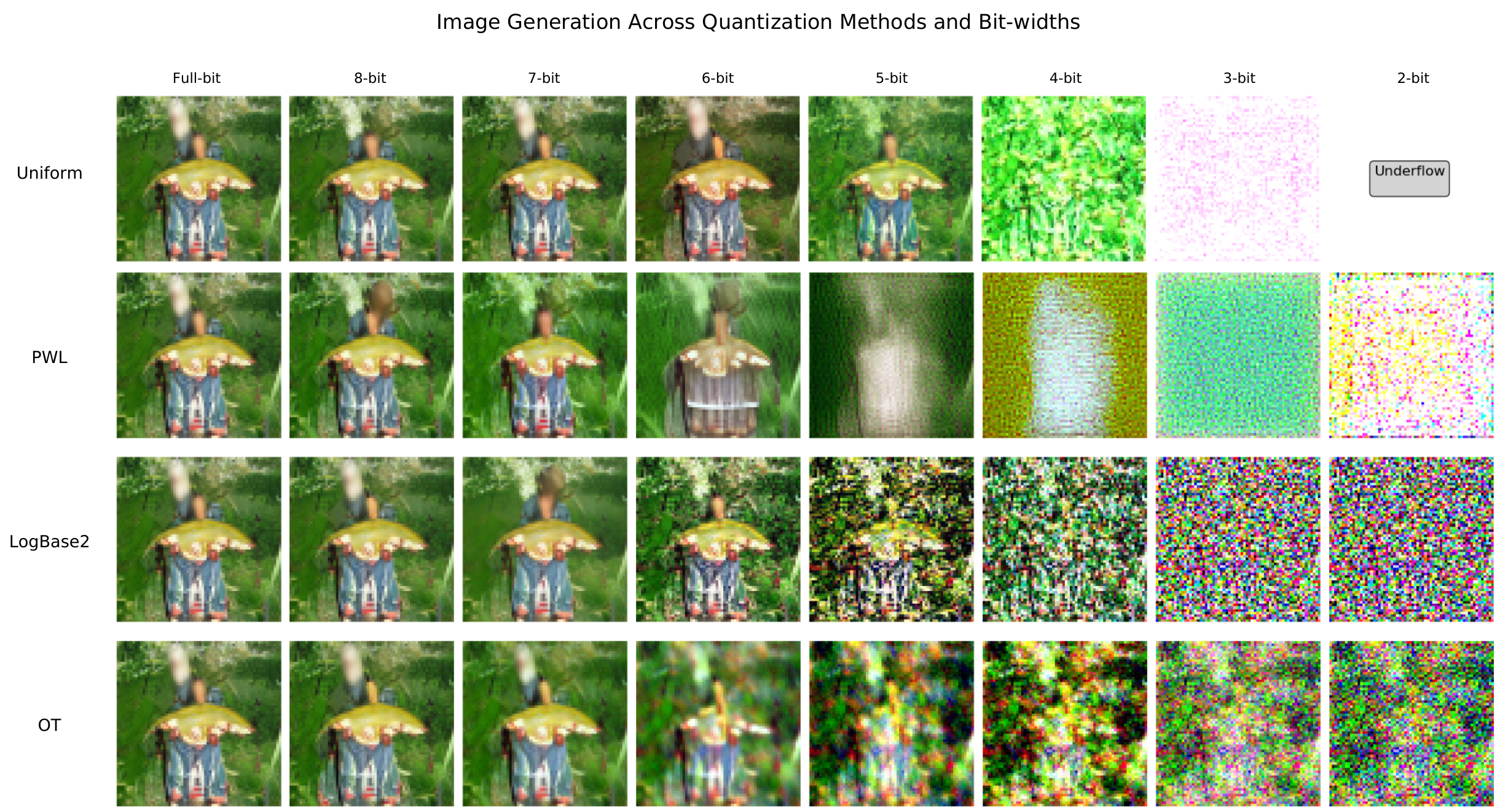}
    \caption{Generated samples for ImageNet.}
    \label{fig:4}
\end{figure}

Generated samples for MNIST, FashionMNIST, CIFAR10, and ImageNet are shown in Figures \ref{fig:1}, \ref{fig:2}, \ref{fig:3}, and \ref{fig:4}, respectively. Across all datasets, the visual fidelity of the samples highlights the stability of our optimal transport quantization pipeline, even as the representation is progressively compressed.

As the bit-width decreases, the quality degrades gracefully rather than abruptly - a behavior that is characteristic of transport-based quantizers but difficult to achieve with conventional uniform or heuristic binning. Remarkably, coherent shapes, class-dependent structure, and dataset-specific textures are preserved down to 2 bits, demonstrating that the learned OT codebook captures semantically meaningful geometry of the latent space rather than merely memorizing high-precision statistics. This robustness at ultra-low precision validates one of the core motivations of our method: that transport-aligned quantization can maintain the generative model’s distributional geometry under aggressive compression.

\end{document}